# From *Submit* to *Submitted* via *Submission*: On Lexical Rules in Large-Scale Lexicon Acquisition.

**Evelyne Viegas, Boyan Onyshkevych**[§]**, Victor Raskin**[§§]**, Sergei Nirenburg**
Computing Research Laboratory,
New Mexico State University,
Las Cruces, NM 88003, USA
`viegas,boyan,raskin,sergei@crl.nmsu.edu`

## Abstract

This paper deals with the discovery, representation, and use of lexical rules (LRs) during large-scale semi-automatic computational lexicon acquisition. The analysis is based on a set of LRs implemented and tested on the basis of Spanish and English business- and finance-related corpora. We show that, though the use of LRs is justified, they do not come cost-free. Semi-automatic output checking is required, even with blocking and preemtion procedures built in. Nevertheless, large-scope LRs are justified because they facilitate the unavoidable process of large-scale semi-automatic lexical acquisition. We also argue that the place of LRs in the computational process is a complex issue.

## 1 Introduction

This paper deals with the discovery, representation, and use of lexical rules (LRs) in the process of large-scale semi-automatic computational lexicon acquisition. LRs are viewed as a means to minimize the need for costly lexicographic heuristics, to reduce the number of lexicon entry types, and generally to make the acquisition process faster and cheaper. The findings reported here have been implemented and tested on the basis of Spanish and English business- and finance-related corpora.

The central idea of our approach – that there are systematic paradigmatic meaning relations between lexical items, such that, given an entry for one such item, other entries can be derived automatically – is certainly not novel. In modern times, it has been reintroduced into linguistic discourse by the Meaning-Text group in their work on lexical functions (see, for instance, (Mel'čuk, 1979).

It has been lately incorporated into computational lexicography in (Atkins, 1991), (Ostler and Atkins, 1992), (Briscoe and Copestake, 1991), (Copestake and Briscoe, 1992), (Briscoe et al., 1993)).

Pustejovsky (Pustejovsky, 1991, 1995) has coined an attractive term to capture these phenomena: one of the declared objectives of his 'generative lexicon' is a departure from sense enumeration to sense derivation with the help of lexical rules. The generative lexicon provides a useful framework for potentially infinite sense modulation in specific contexts (cf. (Leech, 1981), (Cruse, 1986)), due to type coercion (e.g., (Pustejovsky, 1993)) and similar phenomena. Most LRs in the generative lexicon approach, however, have been proposed for small classes of words and explain such grammatical and semantic shifts as $+count$ to $-count$ or $-common$ to $+common$.

While shifts and modulations are important, we find that the main significance of LRs is their promise to aid the task of *massive* lexical acquisition.

Section 2 below outlines the nature of LRs in our approach and their status in the computational process. Section 3 presents a fully implemented case study, the morpho-semantic LRs. Section 4 briefly reviews the cost factors associated with LRs; the argument in it is based on another case study, the adjective-related LRs, which is especialy instructive since it may mislead one into thinking that LRs are unconditionally beneficial.

## 2 Nature of Lexical Rules

### 2.1 Ontological-Semantic Background

Our approach to NLP can be characterized as ontology-driven semantics (see, e.g., (Nirenburg and Levin, 1992)). The lexicon for which our LRs are introduced is intended to support the computational specification and use of text meaning representations. The lexical entries are quite complex, as they must contain many different types of lexical knowledge that may be used by specialist processes for automatic text analysis or generation (see, e.g.,

(Onyshkevych and Nirenburg, 1995), for a detailed description). The acquisition of such a lexicon, with or without the assistance of LRs, involves a substantial investment of time and resources. The meaning of a lexical entry is encoded in a (lexical) semantic representation language (see, e.g., (Nirenburg et al., 1992)) whose primitives are predominantly terms in an independently motivated world model, or ontology (see, e.g., (Carlson and Nirenburg, 1990) and (Mahesh and Nirenburg, 1995)).

The basic unit of the lexicon is a 'superentry,' one for each citation form holds, irrespective of its lexical class. Word senses are called 'entries.' The LR processor applies to all the word senses for a given superentry. For example, *pronunciar* has (at least) two entries (one could be translated as "articulate" and one as "declare"); the LR generator, when applied to the superentry, would produce (among others) two forms of *pronunciación*, derived from each of those two senses/entries.

The nature of the links in the lexicon to the ontology is critical to the entire issue of LRs. Representations of lexical meaning may be defined in terms of any number of ontological primitives, called *concepts*. Any of the concepts in the ontology may be used (singly or in combination) in a lexical meaning representation.

No necessary correlation is expected between syntactic category and properties and semantic or ontological classification and properties (and here we definitely part company with syntax-driven semantics–see, for example, (Levin, 1992), (Dorr, 1993) –pretty much along the lines established in (Nirenburg and Levin, 1992). For example, although meanings of many verbs are represented through reference to ontological EVENTs and a number of nouns are represented by concepts from the OBJECT sublattice, frequently nominal meanings refer to EVENTs and verbal meanings to OBJECTs. Many LRs produce entries in which the syntactic category of the input form is changed; however, in our model, the semantic category is preserved in many of these LRs. For example, the verb *destroy* may be represented by an EVENT, as will the noun *destruction* (naturally, with a different linking in the syntax-semantics interface). Similarly, *destroyer* (as a person) would be represented using the same event with the addition of a HUMAN as a filler of the agent case role. This built-in transcategoriality strongly facilitates applications such as interlingual MT, as it renders vacuous many problems connected with category mismatches (Kameyama et al., 1991) and misalignments or divergences (Dorr, 1995), (Heid, 1993) that plague those paradigms in MT which do not rely on extracting language-neutral text meaning representations. This transcategoriality is supported by LRs.

## 2.2 Approaches to LRs and Their Types

In reviewing the theoretical and computational linguistics literature on LRs, one notices a number of different delimitations of LRs from morphology, syntax, lexicon, and processing. Below we list three parameters which highlight the possible differences among approaches to LRs.

### 2.2.1 Scope of Phenomena

Depending on the paradigm or approach, there are phenomena which may be more–or less–appropriate for treatment by LRs than by syntactic transformations, lexical enumeration, or other mechanisms. LRs offer greater generality and productivity at the expense of overgeneration, i.e., suggesting inappropriate forms which need to be weeded out before actual inclusion in a lexicon. The following phenomena seem to be appropriate for treatment with LRs:

- Inflected Forms - Specifically, those inflectional phenomena which accompany changes in subcategorization frame (passivization, dative alternation, etc.).

- Word Formation - The production of derived forms by LR is illustrated in a case study below, and includes formation of deverbal nominals (*destruction*, *running*), agentive nouns (*catcher*). Typically involving a shift in syntactic category, these LRs are often less productive than inflection-oriented ones. Consequently, derivational LRs are even more prone to overgeneration than inflectional LRs.

- Regular Polysemy - This set of phenomena includes regular polysemies or regular non-metaphoric and non-metonymic alternations such as those described in (Apresjan, 1974), (Pustejovsky, 1991, 1995), (Ostler and Atkins, 1992) and others.

### 2.2.2 When Should LRs Be Applied?

Once LRs are defined in a computational scenario, a decision is required about the time of application of those rules. In a particular system, LRs can be applied at acquisition time, at lexicon load time and at run time.

- Acquisition Time - The major advantage of this strategy is that the results of any LR expansion can be checked by the lexicon acquirer, though at the cost of substantial additional time. Even with the best left-hand side (LHS) conditions (see below), the lexicon acquirer may be flooded by new lexical entries to validate. During the review process, the lexicographer can accept the generated form, reject it as inappropriate, or make minor modifications. If the LR is being used to build the lexicon up from scratch, then mechanisms used by Ostler and Atkins (Ostler and Atkins, 1992) or (Briscoe et al., 1995), such as blocking or preemption, are not available as automatic mechanisms for avoiding overgeneration.

- **Lexicon Load Time** - The LRs can be applied to the base lexicon at the time the lexicon is loaded into the computational system. As with run-time loading, the risk is that overgeneration will cause more degradation in accuracy than the missing (derived) forms if the LRs were not applied in the first place. If the LR inventory approach is used or if the LHS constraints are very good (see below), then the overgeneration penalty is minimized, and the advantage of a large run-time lexicon is combined with efficiency in look-up and disk savings.

- **Run Time** - Application of LRs at run time raises additional difficulties by not supporting an index of all the head forms to be used by the syntactic and semantic processes. For example, if there is an LR which produces *abusive-adj2* from *abuse-v1*, the adjectival form will be unknown to the syntactic parser, and its production would only be triggered by failure recovery mechanisms — if direct lookup failed and the reverse morphological process identified *abuse-v1* as a potential source of the entry needed.

A hybrid scenario of LR use is also plausible, where, for example, LRs apply at acquisition time to produce new lexical entries, but may also be available at run time as an error recovery strategy to attempt generation of a form or word sense not already found in the lexicon.

### 2.2.3 LR Triggering Conditions

For any of the LR application opportunities itemized above, a methodology needs to be developed for the selection of the subset of LRs which are applicable to a given lexical entry (whether base or derived). Otherwise, the LRs will grossly overgenerate, resulting in inappropriate entries, computational inefficiency, and degradation of accuracy. Two approaches suggest themselves.

- **LR Itemization** - The simplest mechanism of rule triggering is to include in each lexicon entry an explicit list of applicable rules. LR application can be chained, so that the rule chains are expanded, either statically, in the specification, or dynamically, at application time. This approach avoids any inappropriate application of the rules (overgeneration), though at the expense of tedious work at lexicon acquisition time. One drawback of this strategy is that if a new LR is added, each lexical entry needs to be revisited and possibly updated.

- **Rule LHS Constraints** - The other approach is to maintain a bank of LRs, and rely on their LHSs to constrain the application of the rules to only the appropriate cases; in practice, however, it is difficult to set up the constraints in such a way as to avoid over- or undergeneration *a priori*. Additionally, this approach (at least, when applied after acquisition time) does not allow explicit ordering of word senses, a practice preferred by many lexicographers to indicate relative frequency or salience; this sort of information can be captured by other mechanisms (e.g., using frequency-of-occurrence statistics). This approach does, however, capture the paradigmatic generalization that is represented by the rule, and simplifies lexical acquisition.

## 3 Morpho-Semantics and Constructive Derivational Morphology: a Transcategorial Approach to Lexical Rules

In this section, we present a case study of LRs based on constructive derivational morphology. Such LRs automatically produce word forms which are polysemous, such as the Spanish *generador* 'generator,' either the artifact or someone who generates. The LRs have been tested in a real world application, involving the semi-automatic acquisition of a Spanish computational lexicon of about 35,000 word senses.

We accelerated the process of lexical acquisition[1] by developing morpho-semantic LRs which, when applied to a lexeme, produced an average of 25 new candidate entries. Figure 1 below illustrates the overall process of generating new entries from a citation form, by applying morpho-semantic LRs.

Generation of new entries usually starts with verbs. Each verb found in the corpora is submitted to the morpho-semantic generator which produces all its morphological derivations and, based on a detailed set of tested heuristics, attaches to each form an appropriate semantic LR label, for instance, the nominal form *comprador* will be among the ones generated from the verb *comprar* and the semantic LR "agent-of" is attached to it. The mechanism of rule application is illustrated below.

The form list generated by the morpho-semantic generator is checked against three MRDs (Collins Spanish-English, Simon and Schuster Spanish-English, and Larousse Spanish) and the forms found in them are submitted to the acquisition process. However, forms not found in the dictionaries are not discarded outright because the MRDs cannot be assumed to be complete and some of these "rejected" forms can, in fact, be found in corpora or in the input text of an application system. This mechanism works because we rely on linguistic clues and therefore our system does not grossly overgenerate candidates.

---
[1]See (Viegas and Nirenburg, 1995) for the details on the acquisition process to build the core Spanish lexicon, and (Viegas and Beale, 1996) for the details on the conceptual and technological tools used to check the quality of the lexicon.

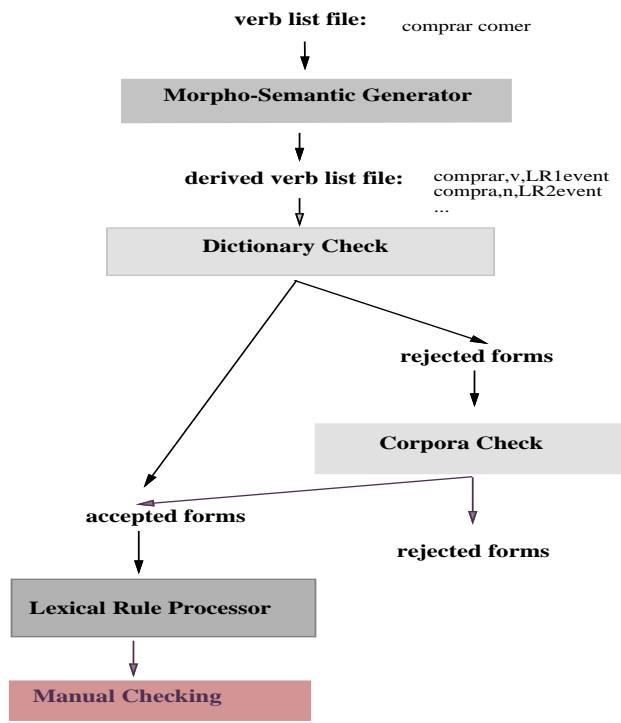

Figure 1: Automatic Generation of New Entries.

Figure 2: Partial Entry for the Spanish lexical item *comprar*.

The Lexical Rule Processor is an engine which produces a new entry from an existing one, such as the new entry *compra* (Figure 3) produced from the verb entry *comprar* (Figure 2) after applying the LR2event rule.[2]

The acquirer must check the definition and enter an example, but the rest of the information is simply retained. The LEXical-RULES zone specifies the morpho-semantic rule which was applied to produce this new entry and the verb it has been applied to.

The morpho-semantic generator produces all predictable morphonological derivations with their morpho-lexico-semantic associations, using three major sources of clues: 1) word-forms with their corresponding morpho-semantic classification; 2) stem alternations and 3) construction mechanisms. The patterns of attachement include unification, concatenation and output rules[3]. For instance *beber* can be derived into *beb[e]dero, bebe[e]dor, beb[i]do, beb[i]da*, *volver* into *vuelto*, and *communicar* into *telecommunicación*, etc... All affixes are assigned semantic features. For instance, the morpho-semantic rule *LRpolarity_negative* is at least attached to all verbs belonging to the -AR class of Spanish verbs, whose initial stem is of the form 'con', 'tra', or 'fir' with the corresponding allomorph -*in* attached to it (*incontrolable, intratable, ...*).

Figure 4 below, shows the derivational morphology output for *comprar*, with the associated lexical rules which are later used to actually generate the entries. Lexical rules[4] were applied to 1056 verb citation forms with 1263 senses among them. The rules helped acquire an average of 25 candidate new entries per verb sense, thus producing a total of 31,680 candidate entries.

From the 26 different citation forms shown in Figure 4, only 9 forms (see Figure 5), featuring 16 new entries, have been accepted after checking.[5]

For instance, *comprable, adj, LR3feasibility-attribute1*, is morphologically derived from *comprar*,

---

[2]We used the typed feature structures (tfs) as described in (Pollard and Sag, 1997). We do not illustrate inheritance of information across partial lexical entries.

[3]The derivation of stem alternations is beyond the scope of this paper, and is discussed in (Viegas et al., 1996).

[4]We developed about a hundred morpho-semantic rules, described in (Viegas et al., 1996).

[5]The results of the derivational morphology program output are checked against existing corpora and dictionaries, automatically.

$$\begin{bmatrix} \text{compra-N1} \\ \text{cat:} & \text{V} \\ \text{dfn:} & \text{acquire the possession or right} \\ & \text{by paying or promising to pay} \\ \text{ex:} \\ \text{admin:} & \text{LR2event "11/12 20:33:02"} \\ \text{syn:} & \left[\text{root: } \boxed{0}\begin{bmatrix}\text{cat: N}\\\text{sem: }\boxed{00}\end{bmatrix}\right] \\ \text{sem:} & [\boxed{00}\text{ buy}] \\ \text{lex-rul:} & \text{comprar-V1 "LR2event"} \\ \ldots \end{bmatrix}$$

Figure 3: Partial Entry for the Spanish lexical item *compra* generated automatically.

and adds to the semantics of *comprar* the shade of meaning of possibility.

In this example no forms rejected by the dictionaries were found in the corpora, and therefore there was no reason to generate these new entries. However, the citation forms *supercompra, precompra, precomprado, autocomprar* actually appeared in other corpora, so that entries for them could be generated automatically at run time.

## 4 The Cost of Lexical Rules

It is clear by now that LRs are most useful in large-scale acquisition. In the process of Spanish acquisition, 20% of all entries were created from scratch by PI-level lexicographers and 80% were generated by LRs and checked by research associates. It should be made equally clear, however, that the use of LRs is not cost-free. Besides the effort of discovering and implementing them, there is also the significant time and effort expenditure on the procedure of semi-automatic checking of the results of the application of LRs to the basic entries, such as those for the verbs.

The shifts and modulations studied in the literature in connection with the LRs and generative lexicon have also been shown to be not problem-free: sometimes the generation processes are blocked–or preempted–for a variety of lexical, semantic and other reasons (see (Ostler and Atkins, 1992)). In fact, the study of blocking processes, their view as systemic rather than just a bunch of exceptions, is by itself an interesting enterprise (see (Briscoe et al., 1995)).

Obviously, similar problems occur in real-life large-scale lexical rules as well. Even the most seemingly regular processes do not typically go through in 100% of all cases. This makes the LR-affected entries not generable fully automatically and this is why each application of an LR to a qualifying phe-nomenon must be checked manually in the process of acquisition.

| Derived form | POS | Lexical Rule |
|---|---|---|
| comprar | v | lr1event |
| compra | n | lr2event8b |
| compra | n | lr2theme_of_event9b |
| comprado | n | lr2reputation_att1a |
| comprador | n | lr2reputation_att2c |
| comprador | n | lr2social_role_rel2c |
| comprado | n | lr2theme_of_event1a |
| comprado | adj | lr3event_telic1a |
| comprable | adj | lr3feasibility_att1 |
| compradero | adj | lr3feasibility_att2c |
| compradizo | adj | lr3feasibility_att3c |
| comprado | adj | lr3reputation_att1a |
| comprador | adj | lr3reputation_att2c |
| comprador | adj | lr3social_role_relc |
| malcomprar | v | neg_eval_attitude1 lr1event |
| malcomprado | adj | lr3event_telic1a |
| subcomprar | v | part_of_relation3 lr1event |
| subcomprado | adj | lr3event_telic1a |
| autocomprar | v | agent_beneficiary1b lr1event |
| autocompra | n | lr2event8b |
| autocompra | n | lr2theme_of_event9b |
| autocomprado | adj | lr3event_telic1a |
| recomprar | v | aspect_iter_semelfact1 lr1event |
| recompra | n | lr2event8b |
| recompra | n | lr2theme_of_event9b |
| recomprado | adj | lr3event_telic1a |
| supercomprar | v | eval_attitude6 lr1event |
| supercompra | n | lr2event8b |
| supercompra | n | lr2theme_of_event9b |
| supercomprado | adj | lr3event_telic1a |
| precomprar | v | before_temporal_rel5 lr1event |
| precompra | n | lr2event8b |
| precompra | n | lr2theme_of_event9b |
| precomprado | adj | lr3event_telic1a |
| descomprar | v | opp_rel2 lr1event |
| descompra | n | lr2event8b |
| descompra | n | lr2theme_of_event9b |
| descomprado | adj | lr3event_telic1a |
| compraventa | n | lr2p_event8b lr2s_event8b |

Figure 4: Morpho-semantic Output.

| Derived form | POS | Lexical Rule |
|---|---|---|
| comprar | v | lr1event |
| comprado | n | lr2theme_of_event1a |
| compra | n | lr2event8b |
| comprado | n | lr2reputation_att1a |
| comprador | n | lr2agent_of2c |
| comprador | n | lr2social_role_rel2c |
| compra | n | lr2theme_of_event9b |
| comprable | adj | lr3feasibility_att1 |
| compradero | adj | lr3feasibility_att2c |
| compradizo | adj | lr3feasibility_att3c |
| comprado | adj | lr3agent_of1a |
| comprador | adj | lr3reputation_att2c |
| comprador | adj | lr3social_role_rel2c |
| comprado | adj | lr3event_telic1a |
| recomprar | v | aspect_iter_semelfact1 lr1event |
| recompra | n | lr2event8b |
| recompra | n | lr2theme_of_event9b |
| compraventa | n | lr2p_event8b lr2s_event8b |

Figure 5: Dictionary Checking Output.

Adjectives provide a good case study for that. The acquisition of adjectives in general (see (Raskin and Nirenburg, 1995)) results in the discovery and application of several large-scope lexical rules, and it appears that no exceptions should be expected. Table 1 illustrates examples of LRs discovered and used in adjective entries.

The first three and the last rule are truly large-scope rules. Out of these, the *-able* rule seems to be the most homogeneous and 'error-proof.' Around 300 English adjectives out of the 6,000 or so, which occur in the intersection of LDOCE and the 1987-89 Wall Street Journal corpora, end in *-able*.

About 87% of all the *-able* adjectives are like *readable*: they mean, basically, something that can be read. In other words, they typically modify the noun which is the theme (or beneficiary, if animate) of the verb from which the adjective is derived:

*One can read the book.–The book is readable.*

The temptation to mark all the verbs as capable of assuming the suffix *-able* (or *-ible*) and forming adjectives with this type of meaning is strong, but it cannot be done because of various forms of blocking or preemption. Verbs like *kill, relate*, or *necessitate* do not form such adjectives comfortably or at all. Adjectives like *audible* or *legible* do conform to the formula above, but they are derived, as it were, from suppletive verbs, *hear* and *read*, respectively. More distressingly, however, a complete acquisition process for these adjectives uncovers 17 different combinations of semantic roles for the nouns modified by the *-ble* adjectives, involving, besides the "standard" theme or beneficiary roles, the agent, experiencer, location, and even the entire event expressed by the verb. It is true that some of these combinations are extremely rare (e.g. *perishable*), and all together they account for under 40 adjectives. The point remains, however, that each case has to be checked manually (well, semi-automatically, because the same tools that we have developed for acquisition are used in checking), so that the exact meaning of the derived adjective with regard to that of the verb itself is determined. It turns out also that, for a polysemous verb, the adjective does not necessarily inherit all its meanings (e.g., *perishable* again).

## 5 Conclusion

In this paper, we have discussed several aspects of the discovery, representation, and implementation of LRs, where, we believe, they count, namely, in the actual process of developing a realistic-size, real-life NLP system. Our LRs tend to be large-scope rules, which saves us a lot of time and effort on massive lexical acquisition.

Research reported in this paper has exhibited a finer grain size of description of morphemic semantics by recognizing more meaning components of non-root morphemes than usually acknowledged.

The reported research concentrated on lexical rules for derivational morphology. The same mechanism has been shown, in small-scale experiments, to work for other kinds of lexical regularities, notably cases of regular polysemy (e.g., (Ostler and Atkins, 1992), (Apresjan, 1974)).

Our treatment of transcategoriality allows for a lexicon superentry to contain senses which are not simply enumerated. The set of entries in a superentry can be seen as an hierarchy of a few "original" senses and a number of senses derived from them according to well-defined rules. Thus, the argument between the sense-enumeration and sense-derivation schools in computational lexicography may be shown to be of less importance than suggested by recent literature.

Our lexical rules are quite different from the lexical rules used in lexically-based grammars (such as (GPSG, (Gazdar et al., 1985) or sign-based theories (HPSG, (Pollard and Sag, 1987)), as the latter can rather be viewed as linking rules and often deal with issues such as subcategorization.

The issue of when to apply the lexical rules in a computational environment is relatively new. More studies must be made to determine the most beneficial place of LRs in a computational process.

Finally, it is also clear that each LR comes at a certain human-labor and computational expense, and if the applicability, or "payload," of a rule is limited, its use may not be worth the extra effort. We cannot say at this point that LRs provide any advantages in computation or quality of the deliverables. What we do know is that, when used justifiably and maintained at a large scope, they facilitate tremendously the costly but unavoidable process of semi-automatic lexical acquisition.

| LRs | Applied to | Entry Type 1 | Entry Type 2 | Examples |
|---|---|---|---|---|
| Comparative | All scalars | Positive Degree | Comparative Degree | good–better big–bigger |
| Semantic Role Shifter Family of LR's | Event-Based Adjs | Adj. Entry corresponding to one semantic role of the underlying verb | Adj. entry corresponding to another semantic role of the underlying verb | abusive noticeable |
| -Able LR | Event-Based Adjs | Verbs taking the -able suffix to form an adj | Adjs formed with the help of -able from these verbs (including "suppletivism") | noticeable vulnerable |
| Human Organs LR | Size adjs | Adjs denoting general human size | Adjs denoting the corresponding size of all or some external organs | undersized-1-2 buxom-1-2 |
| Size Importance LR | Size adjs | Basic size adjs | Figurative meanings of same adjectives | big-1-2 |
| -Scaled LR | VeryTrueScalars (age, size, price,) | True scalar adjectives | Adj-scale(d) | modest-modest(ly)--price(d)old-old-age |
| Negative LR | All adjs | Positive adjs | Corresponding Negative adjectives | noticeable unnoticeable |

Table 1: Lexical Rules for Adjectives.

## 6 Acknowledgements

This work has been supported in part by Department of Defense under contract number MDA-904-92-C-5189. We would like to thank Margarita Gonzales and Jeff Longwell for their help and implementation of the work reported here. We are also grateful to anonymous reviewers and the Mikrokosmos team from CRL.